\documentclass[smallcondensed]{svjour3}
\usepackage{url}
\usepackage{graphicx}
\usepackage{xcolor}
\usepackage{theapa}

\smartqed

\newenvironment{packed_enumerate}{
\begin{enumerate}
  \setlength{\itemsep}{1pt}
  \setlength{\parskip}{0pt}
  \setlength{\parsep}{0pt}
}{\end{enumerate}}

\newcommand{\satzilla}{SATzilla}

\begin{document}
\title{Simple Algorithm Portfolio for SAT}
\date{}
\author{Mladen Nikoli\'c \and Filip Mari\'c \and Predrag Jani\v ci\'c}
\institute{M. Nikoli\'c \at
            Faculty of Mathematics, University of Blegrade, Belgrade, Serbia\\
            \email{nikolic@matf.bg.ac.rs}\\
            Tel.: +381648650064
           \and
           F. Mari\'c \at
            Faculty of Mathematics, University of Blegrade, Belgrade, Serbia\\
            \email{filip@matf.bg.ac.rs}
           \and
           P. Jani\v ci\'c \at
            Faculty of Mathematics, University of Blegrade, Belgrade, Serbia\\
            \email{janicic@matf.bg.ac.rs}
}
\date{Received: date / Accepted: date}

\maketitle

\begin{abstract}
  The importance of algorithm portfolio techniques for SAT has long
  been noted, and a number of very successful systems have been
  devised, including the most successful one --- \satzilla. However,
  all these systems are quite complex (to understand, reimplement,
  or modify). In this paper we present an algorithm portfolio for
  SAT that is extremely simple, but in the same time so efficient that
  it outperforms \satzilla. For a new SAT instance to
  be solved, our portfolio finds its $k$-nearest neighbors from the
  training set and invokes a solver that performs the best for those
  instances. The main distinguishing feature of our algorithm
  portfolio is the locality of the selection procedure --- the
  selection of a SAT solver is based only on few instances similar to
  the input one. An open source tool that implements our approach
  is publicly available.
\end{abstract}


\section{Introduction}
\label{sec:introduction}

Solving time for a SAT instance can significantly vary for different
solvers. Therefore, for many SAT instances, availability of different
solvers may be beneficial. This observation leads to
algorithm portfolios which, among several available solvers, select
one that is expected to perform best on a given instance. This selection
is based on data about the performance of available solvers on a large
training set of instances. The problem of algorithm portfolio is not
limited only to the SAT problem, but can be considered in general
\cite{Huberman97,Gomes01,Horovitz01}.

There are a number of approaches to algorithm portfolios for SAT, the
most successful one being \satzilla{} \cite{Xu08} that regularly wins
in various categories of SAT Competitions\footnote{\url{http://www.satcompetitions.org}}.
SATzilla is very successful, but is a rather complex machinery not
easy to understand, reimplement or modify. In this paper we present an
algorithm portfolio system, based on the $k$-nearest
neighbors method, that is conceptually significantly simpler and
more efficient than SATzilla. It derives from our earlier
research on solver policy selection \cite{Nikolic09}.

The rest of the paper is organized as follows. In Section
\ref{sec:related_work}, some of the existing algorithm portfolios are
described. In Section \ref{sec:algorithm_selection}, the proposed
technique is described and in Section \ref{sec:evaluation} the
experimental results are presented.  The conclusions are drawn in
Section \ref{sec:conclusions}.

\section{Algorithm Portfolios for SAT}
\label{sec:related_work}

Various approaches to algorithm portfolio for SAT and related problems
have been devised \cite{Gomes01,Horovitz01,Lagoudakis01,Samulowitz07},
but the turning point in the field has been marked by the appearance
of \satzilla{} portfolio \cite{Nudelman04,Xu08}. Here we describe
several recent relevant approaches for algorithm selection for SAT,
most of them using fragments of \satzilla{} methodology.

\paragraph{\satzilla.}
\satzilla, the algorithm portfolio that has been dominating recent SAT
Competitions, is the most important and the most successful algorithm
portfolio for SAT, with admirable performance
\cite{Xu08,Xu09}. \satzilla{} represents instances by using different
\emph{features} and then predicts runtime of its constituent solvers
based on these features and relying on \emph{empirical hardness
  models} obtained during the training phase.

\satzilla{} is a very complex system. On a given input instance,
\satzilla{} first runs two {\em presolvers} for a short amount of time, in
a hope that easy instances will be quickly dispatched.
If an instance is not solved by the presolvers,
its features are computed. Since the feature computation can take too long,
before computing features, the feature computation
time is predicted using empirical hardness models. If the estimate is more
than 2 minutes, a {\em backup solver} is run. Otherwise, using computed
features, a runtime for each component solver is predicted. The solver
predicted to be the best is invoked.
If this solver fails (e.g., if it
crashes or runs out of memory), the next best solver is invoked.

The training data are obtained by measuring the solving time for all
instances from some chosen training set by all solvers from some
chosen set of solvers (using some predetermined cutoff time).
For each category of instances (used in SAT Competitions) ---
random, crafted, and industrial, a separate SATzilla system is built.
For each system, a hierarchical empirical hardness model for each
solver is trained to predict its runtime. This prediction
is obtained by combining runtime predictions of separate conditional
models for satisfiable and for unsatisfiable instances. To enable this,
SATzilla uses an estimator of probability whether the input instance is
satisfiable that is trained using sparse multinomial logistic regression.
Each conditional model is obtained in the following manner. First,
starting from a set of base features, a feature selection step is performed in
order to find features that maximally reduce the model training error. Then,
the pairwise products of the remaining features are added as new
features, and the second round of feature selection is
performed. Finally, the ridge regression model for runtime prediction
is trained using the selected features. From the set of solvers that
have been evaluated on the training data, best solvers are chosen for
the component solvers automatically, using a randomized iterative
procedure. The presolvers and the backup solver are also selected
automatically.

\paragraph{ArgoSmArT.}
ArgoSmArT is a system developed for instance-based selection of
policies for a single SAT solver \cite{Nikolic09}. As a
suitable underlying SAT solver it uses a modular solver ArgoSAT
\cite{Maric09}. ArgoSmArT uses a training set of
SAT instances divided manually in classes of instances of similar
origin (e.g., as in the SAT Competition corpus). Each instance is
represented using (a subset of) the \satzilla{} features. For the
input instance to be solved, the feature values are computed and the
nearest (with respect to some distance measure) neighbor instance from the
training set, belonging to some class $c$ is found. Then, the input
instance is solved using the solver configuration that is known to
perform best on the class $c$.

ArgoSmArT does not deal with solver
tuning and assumes that good configurations for classes are provided
in advance. This approach could be used for selection of policies for other
solvers, too. Moreover, it can be also used as an algorithm portfolio.

\paragraph{ISAC.}
ISAC is a solver configurator that also has the potential to be
applied to the general problem of algorithm portfolio
\cite{Kadioglu10}. It divides a training set in families automatically
using a clustering technique. It is integrated with GGA
\cite{Ansotegui09}, a system capable of finding a good solver
configuration for each family. The instances are represented
using \satzilla{} features, but scaled in the interval [-1,1]. For an
input instance, the features are computed and the nearest center of
available clusters is found. If the distance from the center
of the cluster is less
than some threshold value, the best configuration for that cluster is
used for solving. Otherwise, a configuration that performs the best on
the whole training set is used.

\paragraph{Latent class models.}
Another recent approach promotes use of statistical models of solver
behavior (\emph{latent class models}) for algorithm selection
\cite{Silverthorn10}.  The proposed models try to capture the
dependencies between solvers, problems, and run durations. Each
instance from the training set is solved many
times by each solver
and a model is fit to the outcomes observed in the training phase
using the iterative expectation-maximization algorithm. During the
testing phase, the model is updated based on new outcomes.  The
procedure for algorithm selection chooses a solver and runtime duration
trying to optimize discounted utility measure on the long run. The
authors report that their system is roughly comparable to \satzilla.

\paragraph{Non-model-based portfolios.} This, most recent, approach \cite{Malitsky11}
also relies on k-nearest neighbors, and was independently developed in parallel
with our research. However, the two systems differ in some aspects (which will be
shown to be important). This approach uses a standard Euclidean distance to measure the
distance of neighbors, while each feature has to be scaled to the interval
[0,1] to avoid dependence on order of magnitude of numbers involved. Also,
the feature set is somewhat different from the one we use. The
approach was evaluated on random instances from SAT 2009 competition
and gave better results than SATzilla.

\section{Nearest Neighbor-Based Algorithm Portfolio}
\label{sec:algorithm_selection}

The existing portfolio systems for SAT build their models (e.g.,
runtime prediction models, grouping of instances, etc.) in advance,
regardless of characteristics of the input instance. We expected that
a finer algorithm selection might be achievable if a local,
input-specific model is built and used. A simple model of that sort
can be obtained by the $k$-nearest neighbor method \cite{Duda01}, from
just few instances similar to the instance being solved. In the rest
of this section we describe our algorithm portfolio for SAT.

It is assumed that a \emph{training set} of instances is solved by all
solvers from the portfolio, and that the solving times
(within a given cutoff time) are available. Based on these
solving times, for each solver a {\em penalty} can be calculated for
any instance (the greater the solving time, the greater the
penalty). Each instance is represented by a vector of {\em features}.

Our algorithm selection technique is given in Figure
\ref{fig:knnselection}. Basically, \emph{for a new instance to be
  solved, its $k$-nearest neighbors from the training set (with
  respect to some {\em distance measure}) are found, and the solver
  with the minimal penalty for those instances is invoked}. In the
case of ties among several solvers, one of them that performs the best
on the whole training set can be chosen.\footnote{In practice, it is
  highly unlikely that there are more than one such solver, but for
  completeness we allow for such possibility (step 5 of the
  procedure).}

\begin{figure}[t!]
\small
{\sffamily
$S$: Set of available solvers

$T$: Set of feature vectors and solving times for each training instance

$k$: Number of neighbors to be considered ($k\leq|T|$)

$i$: Input instance

\begin{packed_enumerate}
 \item[(Solver selection)\hspace{-22mm}]
 \item $f\leftarrow $\verb|features|$(i)$
 \item $T'\leftarrow$ set of $k$ instances from $T$ nearest to $f$
 \item $S'\leftarrow \{s\in S\ |\ $\verb|penalty|$(s,T')=min_{s'\in S}$\verb|penalty|$(s',T')\}$
 \vspace{2mm}
 \item[(Resolution of ties among solvers from $S'$)\hspace{-59mm}]
 \item $S^*\leftarrow \{s\in S'\ |\ $\verb|penalty|$(s,T)=min_{s'\in S'}$\verb|penalty|$(s',T)\}$
 \vspace{2mm}
 \item[(Solving)\hspace{-10mm}]
 \item Solve $i$ using any $s\in S^*$
\end{packed_enumerate}
}
\caption{$k$-nearest neighbors algorithm portfolio for SAT.}
\label{fig:knnselection}
\end{figure}

To make the method concrete, the set of features, the penalty and the
distance measure have to be defined.

\paragraph{Features.} The authors of \satzilla{} introduced 96
features that are used to characterize SAT instance \cite{Xu08,Xu09},
used subsequently also by other systems \cite{Nikolic09,Kadioglu10}.
The main problem with using a full set of these features is the time
needed to compute them for large instances.\footnote{As said, \satzilla{}
even performs a feature computation time prediction and does the computation
itself only if the predicted time does not exceed 2 minutes.} The features we
chose, given in Figure \ref{fig:features}, can be computed very
quickly. They are some of the purely syntactical ones used by
\satzilla. Though this subset may not be enough for good runtime
prediction that \satzilla{} is based on, it may
serve well for algorithm selection.

\begin{figure}[t!]
\centering
\begin{minipage}{0.8\linewidth}
\centering
{\small
\noindent \textbf{Problem Size Features}:
\begin{packed_enumerate}
\item[1-3.] \textit{Number of clauses} $c$, \textit{Number of variables} $v$, \textit{Ratio} $v/c$
\end{packed_enumerate}

\noindent \textbf{Variable-Clause Graph Features}:
\begin{packed_enumerate}
\item[4-8.] \textit{Variable nodes degree statistics}: mean, variation
  coefficient, min, max, and entropy.
\item[9-13.] \textit{Clause nodes degree statistics}: mean, variation
  coefficient, min, max, and entropy.
\end{packed_enumerate}

\noindent \textbf{Balance Features}:
\begin{packed_enumerate}
\item[14-16.] \textit{Ratio of positive and negative literals in each
    clause}: mean, variation coefficient, and entropy.
\item[17-21.] \textit{Ratio of positive and negative occurrences each
    variable}: mean, variation coefficient, min, max and entropy.
\item[22-23.] \textit{Fraction of binary and ternary clauses}.
\end{packed_enumerate}

\noindent \textbf{Proximity to Horn Formula}:
\begin{packed_enumerate}
\item[24.] \textit{Fraction of Horn clauses}
\item[25-29.] \textit{Number of occurrences in a Horn clause for each
    variable}: mean, variation coefficient, min, max, and entropy.
\end{packed_enumerate}
}
\end{minipage}
\caption{\satzilla{} features used.}
\label{fig:features}
\end{figure}

\paragraph{Penalty.} If a solving time for a solver and for a given
instance is less that a given cutoff time, the penalty for the solver
on that instance is the solving time.  If it is greater then the
cutoff time, for the penalty time we take 10 times the cutoff
time. This is the {\em PAR10 score} \cite{Hutter09}. We define a PAR10
score of a solver on a set of instances to be the sum of its PAR10
scores on individual instances.

\paragraph{Distance measure.} We prefer the distance measure that
performed well for ArgoSmArT:
$$d(x,y)=\sum_i\frac{|x_i-y_i|}{\sqrt{x_iy_i}+1}$$
where $x_i$ and $y_i$ are coordinates of vectors $x$ and $y$
(containing feature values of the instance), respectively.
However, any distance measure could be used.

\bigskip

Compared to the approaches described in Section
\ref{sec:related_work}, our procedure does not discriminate between
satisfiable and unsatisfiable or between random, crafted, and
industrial instances. The procedure does not use presolvers, does not
predict feature computation time, nor it uses any feature selection or
feature generation mechanisms. It is not assumed that the structure of
instance families is given in advance, nor it is constructed in any
way. Also, the algorithm does not use any advanced statistical
techniques, nor does solve the same instances several times.
Compared to the approach of Malitsky et al. \cite{Malitsky11}, we use a
smaller feature set, different distance measure and avoid feature scaling.

Note that the special case of $1$-nearest neighbor technique, has some
advantages compared to the general case. Apart for simpler
implementation, it can have a wider range of application. In the case
of algorithm configuration selection (that can be seen as a special
case of algorithm selection --- each configuration of an algorithm can
be considered as a different algorithm), it would be expensive or
practically impossible to have each instance solved for all algorithm
configurations. Therefore, neither \satzilla{} nor $k$-nearest
neighbor approach for $k>1$ is applicable in this situation.
However, there are special optimization based
techniques for finding good solver configurations off-line
\cite{Hutter09,Ansotegui09}. Hence, for each instance,
one good configuration can be known. This is sufficient for the
$1$-nearest neighbor approach to be used.

\section{Implementation and Experimental Evaluation}
\label{sec:evaluation}

Our implementation of the presented algorithm portfolio for SAT,
ArgoSmArT k-NN,\footnote{The source code and the binary are available from http://argo.matf.bg.ac.rs/ download section.}
consists of less than 2000 lines of C++ code. The core part, concerned
with the solver selection, has around 100 lines of code, while the
rest is a feature computation code, solver initialization and
invocation, time measurement, etc. All the
auxiliary code (everything except the solver selection mechanism) is
derived from the \satzilla{} source code by removing or simplifying
parts of its code.

In the evaluation we compare ArgoSmArT k-NN with SATzilla 2009.
We are not aware of a publicly available implementation related to
the approach of Malitsky at al., but we compare different decisions in our two
approaches within ArgoSmArT k-NN.

Instead of solving instances from a training set, the
training data for \satzilla{} 2009\footnote{\satzilla{} 2009 is a winner of SAT Competition 2009 in random category.}, available from the \satzilla{} web
site\footnote{\url{http://www.cs.ubc.ca/labs/beta/Projects/SATzilla/}}, was used.
\satzilla{} was trained using 5883 instances from SAT Competitions
(2002-2005 and 2007) and SAT Races 2006 and 2008 \cite{Xu09}. The data
available on the web site include the solving information for 4701
instances (the solving data for other instances SATzilla is trained on is not available on the web).
The available solving times of these instances were used,
while the feature values were recomputed (in order to avoid using
the SatELite preprocessor that SATzilla 2009 uses as a first step of
feature computation). The cutoff time of 1500s was
used. When instances that where not solved by any solver within that
time limit and when duplicate instances were excluded, there were 4276
remaining instances in the training set.
The feature vectors of training instances and
their solving times for all solvers used are included in the
implementation of ArgoSmArT (1.3Mb of data).
As a test set, we used all the instances from the SAT Competition 2009.

There are 13 solvers for which the solving data are available, and
that are used by \satzilla{} 2009 as components solvers in 3 versions
of \satzilla{} (random, crafted, and industrial) \cite{Xu09}.
Each version of \satzilla{} uses a specialized subset of these 13
solvers that is detected to perform the best on each category of
instances. No specialized versions of ArgoSmArT are made,
but simply all 13 solvers are used as component solvers.

\begin{figure}
\begin{center}
\includegraphics[width=0.75\textwidth]{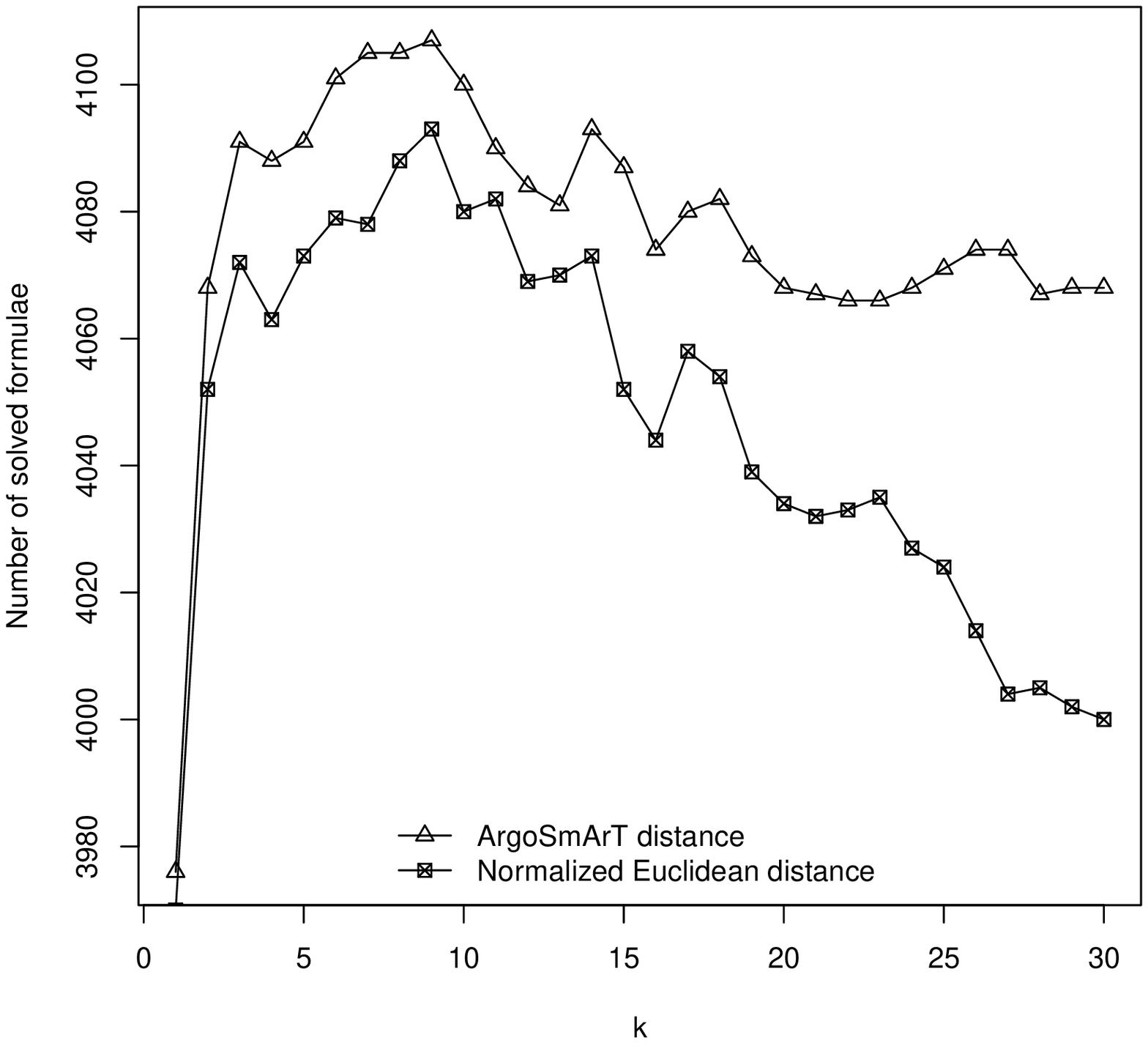}
\caption{The number of solved instances from the training set for ArgoSmArT using each of the compared
distances for each k from 1 to 30.}
\label{fig:compdist}
\end{center}
\end{figure}

ArgoSmArT k-NN can use any distance
measure, but we take into consideration two measures.
The first is the one shown in Section \ref{sec:algorithm_selection}
that performed best for ArgoSmArT, and for some other problems
\cite{Tomovic06}. The second one is the Euclidean distance
(with features being scaled to [0,1]) as used by Malitsky et al.
The comparison of these distances for various $k$ is shown in Figure \ref{fig:compdist}.
The number of solved instances for each distance and each $k$ is obtained by
\emph{leave one out} procedure (the solver to be used for each instance is
chosen by excluding the instance from the training set, and applying
the solver selection procedure using the rest of the training set).
It is obvious that ArgoSmArT distance is uniformly better than the
Euclidean one. For both distances, the highest number of solved instances
is obtained for $k=9$. Hence, we use these choices for ArgoSmArT k-NN
in further evaluation. Also, we justify our choice of features by
measuring the feature computation times. For the full set of 48 features
used by Malitsky et al., minimal, average and maximal computation times on the training set
are 0.002, 19.2 and 6257 seconds. For our, reduced, set of features the minimal,
average and maximal computation times are 0.0017, 0.45, and 51.2 seconds.

An experimental comparison between ArgoSmArT k-NN and any individual
version of \satzilla{} would not be fair since each version is
designed specifically for one kind of instances. So, in order to make
a fair comparison, on random instances we used \satzilla{} random, on
crafted instances we used \satzilla{} crafted, and on industrial
instances we used \satzilla{} industrial. This virtual \satzilla{}
system will be just referred to as \satzilla. In our experimental
comparison, we included MXC08 (the best single solver on the training
set), \satzilla, the ArgoSmArT system based on \cite{Nikolic09} with
13 \satzilla solvers instead of ArgoSAT configurations,
ArgoSmArT 1-NN, and ArgoSmArT 9-NN. Also, we compare to the virtual best
solver --- a virtual solver that solves each instance by the fastest
available solver for that instance (showing the upper achievable
limit). Experiments were performed on a cluster with 32 dual core 2GHz
Intel Xeon processors with 2GB RAM per processor. The results are
given in Table \ref{tab:comp2009} and they show that ArgoSmArT
1-NN/ArgoSmArT 9-NN outperformed all other solvers/portfolios in all categories.

\begin{table}
\centering
\begin{tabular}{ccccccc}
\hline
& MXC08 & SATzilla & ArgoSmArT & ArgoSmArT 1-NN & ArgoSmArT 9-NN & VBS\\
\hline
$Time_{ALL}$ & $>$1500s & 635s & 874s & 390s & 353s & 115s\\
$N_{ALL}$ & \textbf{355} & \textbf{635} & \textbf{609} & \textbf{685} & \textbf{692} & \textbf{816}\\
$N_{RND}$ & 84 & 375 & 308 & 367 & 390 & 454\\
$N_{CRF}$ & 124 & 128 & 154 & 158 & 149 & 188\\
$N_{IND}$ & 147 & 132 & 147 & 160 & 153 & 174\\
\hline
\end{tabular}
\caption{Experimental results on instances from SAT Competition 2009.
For each solver/portfolio the number of solved instances and the median
solving time are given for the whole corpus. Also, the number of solved instances
is given for each of the categories of instances --- random, crafted,
and industrial. The total number of instances is 1143.}
\label{tab:comp2009}
\end{table}

It is a common practice on SAT Competitions and SAT Races to repeat
instances known from previous events. This results in overlapping of
training and test set.  To be thorough, in Table
\ref{tab:comp2009nooverlap}, we provide experimental evaluation on the
test set without the instances contained in the training set.

\begin{table}
\centering
\begin{tabular}{ccccccc}
\hline
& MXC08 & \satzilla & ArgoSmArT & ArgoSmArT 1-NN & ArgoSmArT 9-NN & VBS\\
\hline
$Time_{ALL}$ & $>$1500s & 497s & 895s & 343s & 274s & 76s\\
$N_{ALL}$ & \textbf{243} & \textbf{513} & \textbf{475} & \textbf{533} & \textbf{553} & \textbf{659}\\
$N_{RND}$ & 84 & 375 & 308 & 367 & 390 & 454\\
$N_{CRF}$ & 77 & 68 & 88 & 86 & 83 & 115\\
$N_{IND}$ & 82 & 70 & 79 & 80 & 80 & 90\\
\hline
\end{tabular}
\caption{Experimental results on instances from SAT Competition 2009 without
the instances known from previous SAT Competitions and SAT Races.
For each solver/portfolio the number of solved instances and the median
solving time are given for the whole corpus. Also, the number of solved instances
is given for each of the categories of instances --- random, crafted, and
industrial. The total number of instances is 894.}
\label{tab:comp2009nooverlap}
\end{table}

One can observe that on one subset of instances (industrial instances
that did not appear on earlier competitions), MXC08 component solver
performs better than all the portfolio approaches. This probably
means that the test set is somewhat biased with respect to the
training set. However, the presented results show that ArgoSmArT
k-NN significantly outperformed other entrants on this test set as well.
Possible reasons for this involve two
characteristics of ArgoSmArT k-NN.  First, in contrast to the original
ArgoSmArT and \satzilla, ArgoSmArT k-NN does not use predefined groups
or precomputed prediction models, built regardless of the input
instance. Instead, ArgoSmArT k-NN selects a solver by considering only
a local set of instances similar to the input instance. Second,
ArgoSmArT and \satzilla{} make their choices by considering specific
groups of instances, but these groups are typically large. On the
other hand, ArgoSmArT k-NN considers only a very small number of
instances (e.g., $k=9$) and this eliminates influence of less relevant
instances. Indeed, \satzilla{} improves its predictive performance by
building specific versions for smaller sets of related instances
(i.e., random, crafted, industrial) \cite{Xu08}, which also supports
the above speculation.

\section{Conclusions}
\label{sec:conclusions}

We presented a strikingly simple algorithm portfolio for SAT stemming
from our work on ArgoSmArT \cite{Nikolic09}.
The presented system, ArgoSmArT k-NN, benefits from the \satzilla{}
system in several ways: it uses a subset of \satzilla{} features for
representation of instances, a selection of SAT solvers, \satzilla{}
solving data for the training corpus, and fragments of SATzilla
implementation. However, in its core part --- selection of a solver to
be used --- ArgoSmArT k-NN significantly differs from
\satzilla. Instead of predicting runtimes, our system selects a solver
based on the knowledge about instances from a local neighborhood of
the input instance, using the $k$-nearest neighbors method.  The
proposed system is implemented and publicly available as open source.
The experimental evaluation shows that the particular decisions made in
the design of our system are even better than the decisions made
in the similar recent system \cite{Malitsky11}. Also, it (even the
simplest version ArgoSmArT 1-NN) performed substantially better
than \satzilla --- the most successful, but rather complex, algorithm
portfolio for SAT.  We believe that the presented approach proves there
is a room for improving of algorithm portfolio systems for SAT, not
necessarily by overengineering.

\section*{Acknowledgements}
This work was partially supported by the Serbian Ministry of Science
grant 174021 and by SNF grant SCOPES IZ73Z0\_127979/1. We thank Xu Lin
for clarifying some issues concerning the training data. We thank
Mathematical Institute of Serbian Academy of Sciences and Arts for
allowing us to use their computing resources.

\vskip 0.2in
\bibliographystyle{theapa}
\bibliography{literatura}

\end{document}